\documentclass[10pt,conference,letterpaper]{IEEEtran}

\makeatletter
\def\@IEEEsectpunct{ }
\renewcommand\paragraph{\@startsection{paragraph}{4}{\z@}
  {1.25ex \@plus .5ex \@minus .2ex}
  {-1em}
  {\normalfont\normalsize\bfseries}}
\makeatother

\usepackage{amsmath,amsfonts}
\usepackage{array}
\usepackage[caption=false,font=small]{subfig}
\usepackage{textcomp}
\usepackage{stfloats}
\usepackage{url}
\usepackage{verbatim}
\usepackage{graphicx}
\usepackage{adjustbox}
\usepackage{cite}
\usepackage{times}
\usepackage{microtype}
\usepackage{soul}
\usepackage[hidelinks]{hyperref}
\usepackage[utf8]{inputenc}
\usepackage{amsthm}
\usepackage{booktabs}
\usepackage[switch]{lineno}

\usepackage{boldline}
\usepackage{multirow}
\usepackage{colortbl}
\usepackage[table]{xcolor}
\usepackage[normalem]{ulem}
\useunder{\uline}{\ul}{}
\usepackage{enumitem}
\usepackage{arydshln}
\usepackage{tabularx}
\usepackage[ruled,lined]{algorithm2e}
\usepackage{bm}
\usepackage{stfloats}
\usepackage{amssymb}
\usepackage[dvipsnames]{xcolor}

\usepackage{float}
\usepackage{newfloat}
\usepackage{listings}
\DeclareCaptionStyle{ruled}{labelfont=normalfont,labelsep=colon,strut=off}
\floatstyle{ruled}
\newfloat{listing}{tb}{lst}{}
\floatname{listing}{Listing}

\makeatletter
\g@addto@macro\normalsize{%
  \setlength\abovedisplayskip{4pt plus 1pt minus 2pt}%
  \setlength\belowdisplayskip{4pt plus 1pt minus 2pt}%
  \setlength\abovedisplayshortskip{2pt plus 1pt}%
  \setlength\belowdisplayshortskip{2pt plus 1pt}}
\makeatother

\newcommand{\model}{TSMini}

\hypersetup{
  pdftitle={TSMini: A Simple Yet Highly Effective  Trajectory Similarity Learning Model},
  pdfauthor={Yanchuan Chang, Dingyang Lyu, Xu Cai, Christian S. Jensen, Jianzhong Qi}
}

\begin{document}

\title{\model: A Simple Yet Highly Effective  Trajectory Similarity Learning Model}

\author{
\IEEEauthorblockN{%
Yanchuan Chang\IEEEauthorrefmark{1},
Dingyang Lyu\IEEEauthorrefmark{1},
Xu Cai\IEEEauthorrefmark{2},
Christian S. Jensen\IEEEauthorrefmark{3},
Jianzhong Qi\IEEEauthorrefmark{1}$^{, 1}$}
\IEEEauthorblockA{%
\IEEEauthorrefmark{1}The University of Melbourne, Australia\quad
\IEEEauthorrefmark{2}National University of Singapore, Singapore\quad
\IEEEauthorrefmark{3}Aalborg University, Denmark\\
\{yanchuan.chang, jianzhong.qi\}@unimelb.edu.au,\quad
dingyang.lyu@student.unimelb.edu.au,\quad
caix@u.nus.edu,\quad
csj@cs.aau.dk}
}

\maketitle
\renewcommand{\thefootnote}{1}
\footnotetext{Corresponding author.}

\begin{abstract}
    
Trajectory similarity is fundamental to many spatio-temporal data mining applications. Recent studies propose deep learning models to approximate conventional trajectory similarity measures, exploiting their fast inference time once trained. Although efficient inference has been reported, challenges remain in similarity approximation accuracy due to difficulties in trajectory granularity modeling and in exploiting similarity signals in training data.
To fill this gap, we propose \model, a highly effective trajectory similarity model with a sub-view modeling mechanism and a $k$ nearest neighbor-based loss. The former enables learning multi-granularity trajectory patterns, while the latter guides \model\ to learn not only absolute similarity values between trajectories but also their relative similarity ranks. Together, these  innovations enable highly accurate trajectory similarity approximation. 
Experiments show that \model\ outperforms the state-of-the-art models by 15\% on average when learning widely used trajectory similarity measures.

\end{abstract}


\section{Introduction}\label{sec:introduction}

Trajectory similarity measures quantify the similarity between two trajectories and thus play an essential role in many spatio-temporal data mining tasks and queries, such as trajectory clustering~\cite{edr,representative_traj}, anomaly detection~\cite{laxhammar2013online,liu2020online}, and $k$-nearest neighbor ($k$NN) queries~\cite{dita,torch}. 
Conventional measures~\cite{dfrechet,edr,edwp} (a.k.a. \emph{non-learned measures}) are typically based on heuristic rules that match up the points in two trajectories. The similarity values are then determined by the distances between matched points. 
These measures often use dynamic programming to speed up the computation, which, however, remain costly for long trajectories~\cite{trajsimi_survey}.

Recent studies~\cite{neutraj,gts,t3s,st2vec,trajgat,trajcl,kgts,movsemcl} adopt deep learning models to learn trajectory embeddings to accelerate similarity computation (a.k.a. \emph{learned measures}).
The idea is that trajectory embeddings are pre-computed offline (or computed online once)  and are then used multiple times. Once embeddings are obtained, the similarity between two trajectories can be approximated by the distance between their embeddings, which can be computed efficiently. 

\begin{figure}[t]
    \centering
    \includegraphics[width=0.98\columnwidth]{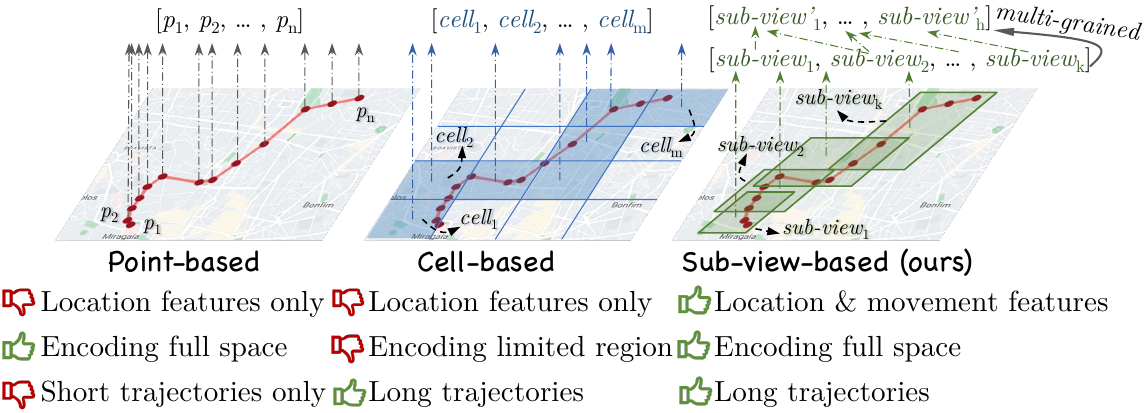}
    \vspace{-2mm}
    \caption{Different forms of trajectory input modeling.}\label{fig:point_modeling}
    \vspace{-1.5mm}
\end{figure}

Despite the high efficiency reported in existing studies~\cite{t3s,tmn,trajcl}, we observe two issues in the learned similarity measures that limit their accuracy:

(1)~\textbf{Difficulties in modeling the movement patterns in a trajectory at different spatial granularities}: 
Existing solutions model an input trajectory as a sequence of points or cells enclosing the trajectory points (see ``point-based'' and ``cell-based'' in Figure~\ref{fig:point_modeling}), or they use both types of inputs together. 
Point-based approaches consume each single raw GPS point and capture locations passed by the trajectories, while they are hard to reflect explicit relationships between points (i.e., movement patterns). 
Cell-based approaches discretize the underlying space with a grid and replace GPS points by the cells enclosing them. The cell-based representation only captures rough locations passed by the trajectories, and it is difficult to determine a grid granularity that retains just the right level of detail of the trajectories. 
Both approaches rely on a single granularity and may not retain the key movement patterns of different trajectories.

(2)~\textbf{Difficulties in fully 
exploiting similarity signals in training data}:
Existing solutions minimize the \emph{mean square error} (MSE) between the predicted similarity of models and the ground-truth similarity, which is typically defined by a non-learned measure.  The ground-truth similarity values can come from a large continuous domain, which may not be fully reflected by the MSE over individual training samples. Thus, it is difficult for a model to learn the similarity concept in the training data from just examining the MSE results.

To address these issues, we propose \model, \emph{a simple yet highly effective trajectory similarity learning model with a sub-view encoder and a $k$NN-guided loss}. 

The \emph{sub-view encoder} prepares and encodes multi-grained sub-views of input trajectories. 
As Figure~\ref{fig:point_modeling} shows,  
a sub-view of a trajectory is a consecutive sub-sequence of the trajectory points that captures fine-grained local movement patterns. 
The decomposition of a trajectory into sub-views is done recursively, forming a series of sub-views capturing  movement patterns at different granularities (cf.~Figure~\ref{fig:subview}).

The output of the sub-view encoder is fed to a one-layer trajectory encoder for trajectory embedding learning. 
This simple encoder is a side benefit of our multi-grained sub-view modeling, since much of the pattern learning task occurs in the sub-view encoder.

To train \model, we employ a \emph{$k$NN-guided loss function} to better exploit training signals in the data. 
The intuition is that, instead of learning from individual similarity values defined by some non-learned measure (e.g., DTW~\cite{dtw}) between each trajectory pair, we examine the relative similarity between different pairs of trajectories to guide \model\ to better learn the similarity space defined by the non-learned measure. 
The $k$NN-guided loss achieves this by imposing a penalty when a trajectory and one of its  $k$NN trajectories are predicted to be less similar than the trajectory and any of its non-$k$NNs.

To sum up, our main contributions are as follows: 
(1)~We propose \model, a simple yet highly effective model for trajectory similarity learning with a sub-view encoder and a $k$NN-guided loss. 
(2)~The sub-view encoder captures multi-grained movement patterns in individual trajectories, while the $k$NN-based loss guides \model\ to learn the  relative similarity among multiple pairs of trajectories. Together, they enable \model\ to better learn the trajectory similarity space. 
(3)~We conduct extensive experiments on three large real  datasets. The results show that \model\ outperforms the state-of-the-art models substantially and consistently, with an accuracy improvement of 15\% on average. 

\section{Related Work}\label{sec:relatedwork}

\subsection{Non-learned Trajectory Similarity Measures}

Non-learned measures generally follow a ``match-then-measure'' paradigm: they match close point pairs and derive the overall similarity from matched-point distances~\cite{edr,erp,edwp,dtw,dfrechet,lcss,hausdorff}.
Most such measures rely on dynamic programming or enumeration under hand-crafted rules, typically incurring quadratic time complexity in the number of trajectory points.

For example, DTW~\cite{dtw} aligns points on two trajectories by minimizing the sum of point-to-point distances, while Discrete Fr\'echet~\cite{dfrechet} measures trajectory similarity based on the maximum distance between the matched points.
Edit-distance-based measures include EDR~\cite{edr}, ERP~\cite{erp}, and EDwP~\cite{edwp}, where EDwP finds optimal point matches by minimizing projected-point costs.
A few measures~\cite{sax_traj1,stsjoin} achieve linear time complexity but require temporal alignment. This paper focuses on spatial similarity measures, while the proposed model can be extended to spatio-temporal measures.

\subsection{Learned Trajectory Similarity Measures}
Learned measures encode trajectories into embeddings for similarity computation. Most methods use a fixed spatial granularity, taking either raw GPS points or discretized cells/road segments as input.
Early RNN-based methods include NEUTRAJ~\cite{neutraj}, which encodes raw points with similarity-weighted MSE, TMN~\cite{tmn}, which predicts pairwise similarity without reusable embeddings, and KGTS~\cite{kgts}, which learns from knowledge-graph cell embeddings with a GRU~\cite{gru}.

Recent self-attention methods include T3S~\cite{t3s}, DTisT~\cite{dtist}, TrajCL~\cite{trajcl}, MovSemCL~\cite{movsemcl}, and TrajGAT~\cite{trajgat}. They encode raw point/cell sequences, fixed-size patches, or quadtree cells~\cite{quadtree}, but still rely on predefined spatial granularity.
MovSemCL~\cite{movsemcl} partitions trajectories into fixed-sized patches and applies self-attention to model both local and long-range patterns.
Its patch granularity is predefined, preventing the method from dynamically adapting the spatial resolution to capture trajectory details or  diverse movement patterns.

TrjSR~\cite{trjsr} and ConvTraj~\cite{Convtraj} are image-based methods that transform trajectories into 2D images and use convolutional layers. Their generated images are tied to specific spatial regions, limiting generalizability, and they learn stand-alone measures rather than approximating a given ground-truth measure.
Other studies target alternative settings: TMLKD~\cite{tmlkd} uses knowledge distillation for label-scarce scenarios, while SimRN~\cite{simrn}, UniTR~\cite{unitr}, and TRACK~\cite{track} assume map-matched trajectories and explicit road-graph modeling. These settings introduce constraints beyond the scope of our study.

Unlike existing methods that encode trajectories at a single or fixed set of granularities, our recursive sub-view encoder constructs multi-grained sub-views from sub-trajectories, retaining both local movement patterns and global structural characteristics.
Moreover, the $k$NN-guided loss leverages relative similarity rankings among multiple trajectory pairs, whereas previous methods typically optimize the mean square error on absolute similarity scores.

\begin{figure*}[t]
    \centering
    \includegraphics[width=0.99\textwidth]{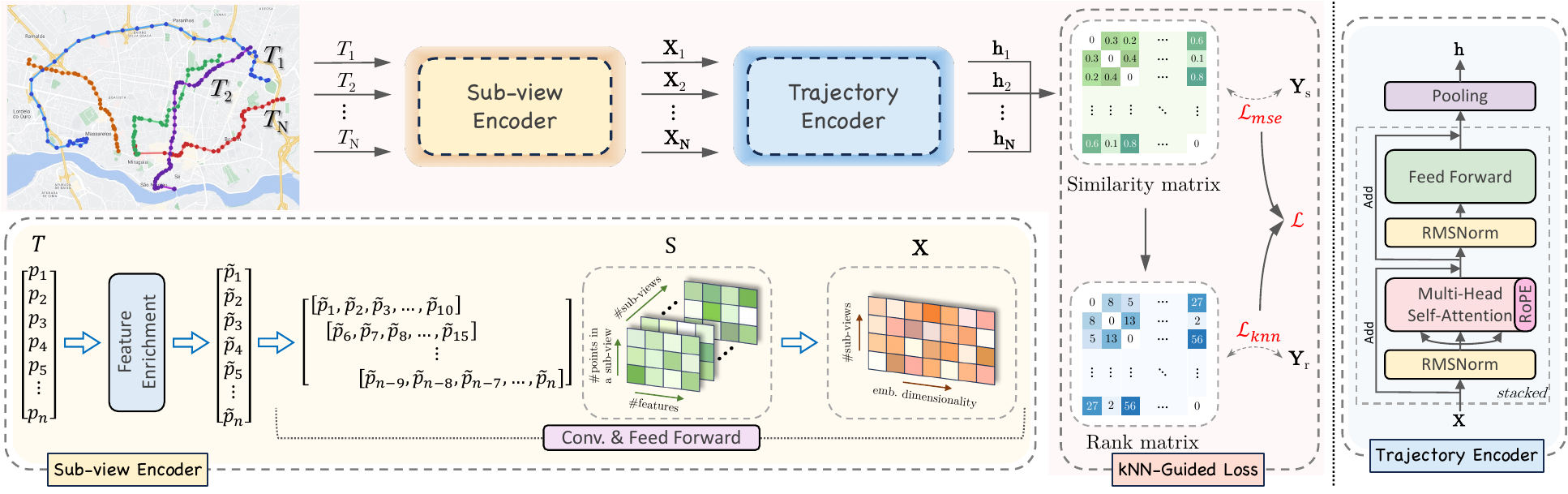}
    \caption{Overview of \model: \model\ first processes raw  points on a trajectory $T$ into multi-grained sub-views and generates sub-view embeddings $\mathbf{X}$ by a Sub-view Encoder. The embeddings are fed into a Trajectory Encoder to generate embeddings $\mathbf{h}$. Pairwise distances between $\mathbf{h}_1, \ldots, \mathbf{h}_N$ of a mini-batch form a predicted \emph{similarity matrix}, which is supervised by the ground-truth similarities $\mathbf{Y}_s$ via the MSE loss $\mathcal{L}_{mse}$. Sorting its rows yields a \emph{rank matrix}, which is supervised by the ground-truth $k$NN ranks $\mathbf{Y}_r$ via our $k$NN-guided loss $\mathcal{L}_{knn}$. Their combination $\mathcal{L}$ lets \model\ learn both exact and relative similarity.}\label{fig:overview}
    \vspace{-2mm}
\end{figure*}

\section{Preliminaries}\label{sec:preliminary}

\paragraph{Trajectory}
A \textit{trajectory} $T= [p_1, p_2, \cdots, p_n]$ is a sequence of $n$ location points, where $p_i$ is the $i$-th point (represented by a pair of coordinates), and
$\overline{p_i p_j}$ is the segment from $p_i$ to $p_j$.

\paragraph{Trajectory Similarity Query}
Given a trajectory dataset $\mathcal{D}$, a query trajectory $T_q$, query parameter $k$, and a trajectory similarity measure $f$, a \textit{trajectory similarity query} (a.k.a. trajectory $k$NN query) returns a set $\mathcal{S} \subseteq \mathcal{D}$ with $|\mathcal{S}| = k$ such that $\forall T\in\mathcal{S}$ and $\forall T'\in\mathcal{D}\setminus \mathcal{S}, f(T, T_q) \geqslant f(T', T_q)$.

\paragraph{Problem Statement}
Given a trajectory similarity measure $f$ and a trajectory dataset $\mathcal{D}$, our aim is to learn a trajectory encoder model $g: T \rightarrow \textbf{h}$, where $\textbf{h} \in \mathbb{R}^d$ is a $d$-dimensional embedding vector, with the following two \emph{Goals}:

(1)~For trajectories $T_i, T_j \in \mathcal{D}$, the difference between the ground-truth similarity value $f(T_i, T_j)$ and the predicted value by the model $f'(T_i, T_j)$ is minimized, where $f'(T_i, T_j) = 1- \text{dist}(g(T_i), g(T_j))$ and
$\text{dist}(\cdot)$ is a simple distance metric such as the $L_1$ or $L_2$ norm.
The ground truth $f(T_i, T_j) = 1 - \text{dist}_f(T_i, T_j)/\text{dist}_{\max} \in [0,1]$ is derived from a non-learned distance $\text{dist}_f$ (e.g., Fr\'echet), with $\text{dist}_{\max}$ the largest pairwise distance in the training set.

(2)~For any $T_i \in \mathcal{D}$ as the query trajectory, the difference between the ground-truth trajectory similarity query result set $\mathcal{S}$ and the computed result set $\mathcal{S}'$ is minimized, 
where $\mathcal{S}'$ is obtained by applying $f'$ as the similarity measure.

\paragraph{Discussion}
Existing studies focus mainly on Goal~1 to minimize the errors in similarity approximation, while such a training goal does not provide sufficient signals for trajectory similarity learning. Based on the similarity values, we further exploit Goal~2 to learn the relative similarity between trajectories for more accurate similarity approximation.

\section{The \model\ Model}\label{sec:method}

Figure~\ref{fig:overview} illustrates the overall model structure.
\model\ first encodes trajectories into sub-view embeddings through a \emph{sub-view encoder} and then  maps the sub-view embeddings into trajectory embeddings through a \emph{trajectory encoder}.
Finally, \model\ uses a \emph{$k$NN-guided loss} for model training.

\subsection{Sub-view Encoder}\label{subsec:method_patch}

Given a raw trajectory input $T$, the sub-view encoder (denoted as \textbf{SVEnc}) first transforms $T$ into sub-sequences,
which we call \emph{sub-views}, instead of modeling individual points as in previous works~\cite{neutraj,t3s,trajcl}.
Trajectory sub-views can capture not only the spatial features of individual points but also local movement patterns. They provide multi-grained spatial features while reducing the input length for the subsequent trajectory encoder, hence leading to embeddings that better preserve trajectory features.
SVEnc augments and encodes the sub-views of trajectory $T$ into embeddings $\mathbf{X} \in \mathbb{R}^{m\times d}$, where $m$ denotes the number of sub-views created on $T$ and $d$ is the embedding dimensionality as before.

\paragraph{Feature Preparation}
Given an input raw trajectory $T$, we augment it by adding spatial features to each point.
A point $p_i \in T$ is expanded to a vector of  seven elements, including the longitude, the latitude, the lengths of segments $\overline{p_{i-1} p_i}$ and $\overline{p_i p_{i+1}}$, the angle between the $x$-axis and $\overline{p_{i-1} p_i}$, the angle between the $x$-axis and $\overline{p_i p_{i+1}}$, and the interior angle between $\overline{p_{i-1} p_i}$ and $\overline{p_i p_{i+1}}$.
We use $\mathbf{S} \in \mathbb{R}^{n\times 7}$ to denote the augmented representation of $T$.

\begin{figure}[t]
    \centering
    \includegraphics[width=0.98\columnwidth]{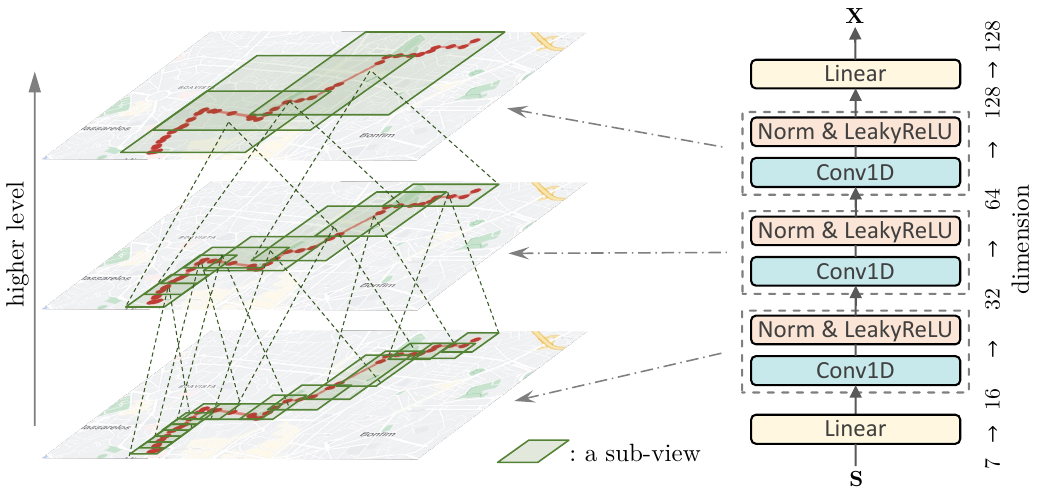}
    \vspace{-3mm}
    \caption{Multi-grained trajectory sub-view modeling.}\label{fig:subview}
    \vspace{-2mm}
\end{figure}

\paragraph{SVEnc}
Figure~\ref{fig:subview} shows the structure of SVEnc.
Inspired by time series representation learning~\cite{patchfy,timellm}, we stack
three convolutional sub-modules sandwiched between two linear layers. Each convolutional sub-module contains a \emph{1D convolutional} (Conv1D) layer, a \emph{batch normalization} layer~\cite{batchnorm}, and a \emph{LeakyReLU} activation function.
For all Conv1Ds, the kernel size $w_{k}$ is set as 3 and the stride $w_{s}$ is set as 1. The dimensionality of each layer is labeled in Figure~\ref{fig:subview}.
Note that we do not explicitly create sub-views. Instead, we feed the entire trajectory $\mathbf{S}$ into SVEnc, where the sub-views are created and encoded simultaneously.
The convolution kernels move  along the dimension of trajectory points, which can be seen as first forming a sequence of sub-views of every $w_{k}$ points and then producing embeddings for the sub-views.
The first Conv1D layer processes sub-views of points, while the second and third Conv1D layers aggregate every $w_{k}$ sub-views from the previous layer, summarizing the movement patterns of longer sub-sequences.
The Conv1Ds use no padding, so each layer shortens the sequence by $w_k-1$, giving $m = n - 3(w_k-1)$ sub-views with receptive fields of 3, 5, and 7 points at the three levels. Positions padded for batching are masked in the subsequent attention.
The output $\mathbf{X} \in \mathbb{R}^{m\times d}$ ($m < n$) of SVEnc  is fed into the trajectory encoder to generate the trajectory embedding $\mathbf{h}$.

\subsection{Trajectory Encoder}\label{subsec:traj_enc}
The trajectory encoder (denoted as \textbf{TrajEnc}) can be any existing sequence embedding model.
Inspired by modern Transformer designs such as Llama-3~\cite{llama3paper}, we use a Transformer encoder layer equipped with RMSNorm, RoPE, and SwiGLU.
This choice comes with two advantages: (1) Self-attention is both effective and efficient for trajectory similarity learning~\cite{t3s,trajsimi_survey}.
(2) RMSNorm, RoPE, and SwiGLU are effective and lightweight components for stabilizing training and capturing sequential correlations beyond vanilla self-attention~\cite{transformer}: RMSNorm stabilizes training on variable-length inputs at lower cost than LayerNorm, RoPE encodes the \emph{relative} order of sub-views, which matters more than absolute positions for matching movement patterns, and SwiGLU adds nonlinearity with fewer parameters than a wider FFN.

Next, we detail TrajEnc, as illustrated by the right part of  Figure~\ref{fig:overview}.
The sub-view embeddings $\mathbf{X}$ first go through a \emph{root mean square normalization} layer (RMSNorm)~\cite{rmsnorm} to obtain normalized input, for more stable model training.
We abuse the notation slightly and also use  $\mathbf{X}$ to denote the normalized input hereafter for conciseness, when there is no ambiguity.

Then, the normalized $\mathbf{X}$ is fed into a \emph{multi-head self-attention} layer  to learn the hidden correlations between trajectory sub-views and generate new embeddings $\mathbf{H}$:
\begin{equation}\label{eq:selfattention}
\begin{split}
    \mathbf{Q}^{j}, \mathbf{K}^{j}, \mathbf{V}^{j} & = \mathbf{X}\mathbf{W}_{Q}^{j}, \mathbf{X}\mathbf{W}_{K}^{j}, \mathbf{X}\mathbf{W}_{V}^{j} \\
    \mathbf{H}^j & = \mathrm{Softmax} (\alpha_{A} \cdot
    \mathrm{RoPE}(\mathbf{Q}^{j}, \mathbf{K}^{j})) \mathbf{V}^{j} \\
    \mathbf{H} & = \mathrm{Concat}(\mathbf{H}^{1},\cdots, \mathbf{H}^{j},\cdots, \mathbf{H}^{h})\mathbf{W}_{H}
\end{split}
\end{equation}
\noindent Here, $\mathbf{W}_Q^j$, $\mathbf{W}_K^j$, and $\mathbf{W}_V^j$ (all in $\mathbb{R}^{d\times (d/h)}$) are learnable weights of the $j$-th self-attention head ($h$ heads in total), $\alpha_{A}$ is a scaling factor, and $\mathbf{W}_{H} \in \mathbb{R}^{d\times d}$. $\mathrm{RoPE}(\cdot)$ computes the attention coefficients and integrates with positional encodings. We adopt the \emph{Rotary Position Embedding} (RoPE)~\cite{rope} for its capability in exploiting the relative position dependency among the sub-views.

Then, we apply a \emph{residual connection}~\cite{resnet} to the output of the self-attention layer and the initial input $\mathbf{X}$ (before normalization), i.e., \emph{adding} the two values. This alleviates gradient vanishing and yields smooth gradients.

Next, the sum from the residual connection is fed into another sub-module that consists of an RMSNorm, a feed forward network (FFN) with an \emph{SwiGLU} activation function~\cite{glu}, and a residual connection, which brings the nonlinear representation learning capability.
The output of this sub-module, $\widetilde{\mathbf{H}}$, has the same shape as $\mathbf{H}$ and $\mathbf{X}$.

Although Transformer encoders may stack multiple layers, we use a single layer to keep TrajEnc compact and efficient.
Finally, we apply \emph{average pooling} to the hidden output $\widetilde{\mathbf{H}}$ to obtain the final trajectory embedding of $T$, i.e., $\mathbf{h} \in \mathbb{R}^{d}$.

\paragraph{Discussion}
TrajEnc is a bidirectional encoder for complete trajectories, not the decoder used in Llama-3.

(1) TrajEnc does \emph{not} adopt \emph{grouped-query attention} (GQA)~\cite{gqa}, which shares attention coefficients among heads to speed up decoder inference at some cost in accuracy. TrajEnc already has few parameters and does not need this trade-off.

(2) TrajEnc does \emph{not} use \emph{causal masking}, which prevents future tokens from affecting attention in autoregressive language models.
This is because an entire trajectory is known for encoding in our task. To encode a  point,
it is more beneficial to consider both preceding and subsequent (i.e., future) points on the trajectory, instead of just the preceding ones.

\subsection{\texorpdfstring{$K$NN-Guided Optimization}{KNN-Guided Optimization}}\label{subsec:method_loss}

Next, we present the $k$NN-guided loss.
To simplify the symbols, we abuse the notation slightly and use the same symbol to represent both a random variable and an observed value of the variable, e.g., $\mathbf{x}$, when the context provides clarity.

\begin{figure}[th]
    \centering
    \includegraphics[width=0.98\columnwidth]{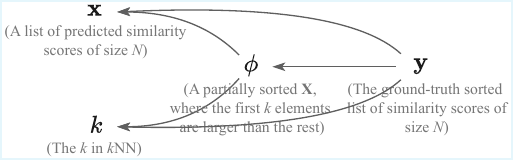}
    \vspace{-3mm}
    \caption{Variable dependency for the $k$NN-guided loss.}\label{fig:probability}
    \vspace{-2mm}
\end{figure}

\paragraph{$K$NN-guided Loss}
We formulate the $k$NN-guided loss $\mathcal{L}_{knn}$ in a probabilistic manner---see Figure~\ref{fig:probability}.
The training set contains $N$ trajectories, and we use each trajectory once as the query trajectory, yielding $N$ training instances.
Let $\mathbf{x}$ be a given variable for a list of size $N$ of the predicted similarity scores between a trajectory and all trajectories in the training set.
Let $k \in \mathcal{K}$ be the query parameter $k$ of $k$NNs, where $\mathcal{K} = \mathbb{N}_{\leq N}$.
We introduce $\phi \in \Phi$  as a hidden variable determined by $\mathbf{x}$ and $k$.
Here, $\phi$ denotes a partially sorted list of size $N$, where the first $k$ elements are larger than the rest, i.e., the first $k$ elements in $\phi$ form the $k$NN result set.
We use $\mathbf{y}$ to represent the ground-truth list of similarity scores of size $N$, which are impacted by the input variables $\mathbf{x}$, $k$, and $\phi$.
The likelihood function for $\mathbf{y}$ conditioned on $\mathbf{x}$ is:
\vspace{-1mm}
\begin{equation}\label{eq:knnloss_1}
    P(\mathbf{y} | \mathbf{x}) =  \sum_{k \in \mathcal{K}} P(\mathbf{y} | \mathbf{x}, k) P(k)
\end{equation}
\vspace{-1mm}
Given a set of $N$ training instances $\{(x, y)\}$ (recall that we use each of the $N$ training trajectories once as the query trajectory, together forming $N$ training instances),
where $x$ is a list of predicted similarity scores between a trajectory and all trajectories in the training set, and $y$ is the ground-truth list,
the $k$NN-guided loss $\mathcal{L}_{knn} :(\mathbf{y}, \mathbf{x}) \rightarrow \mathbb{R}$ is designed to minimize the negative log-likelihood, as shown in Eq.~\ref{eq:knnloss_1}:
\vspace{-1mm}
\begin{equation}\label{eq:knnloss_2}
    \mathcal{L}_{knn}(\mathbf{y}, \mathbf{x}) = -\log_2 \sum_{k \in \mathcal{K}} P(\mathbf{y} | \mathbf{x}, k) P(k)
\end{equation}
\vspace{-1mm}
This log-sum loss is difficult to optimize directly because, when the summed probability stays close to 1, the loss remains near 0 and may yield vanishing gradients. Since the negative logarithm is convex, Jensen's inequality gives a surrogate upper bound for Eq.~\ref{eq:knnloss_2}. Further, since $P(k)$ always lies between 0 and 1, the bound simplifies to Eq.~\ref{eq:knnloss_2_2} and is denote as $\mathcal{L}_{knn}^{+}$:
\vspace{-1mm}
\begin{align}
    \mathcal{L}_{knn}(\mathbf{y}, \mathbf{x}) & = -\log_2 \sum_{k \in \mathcal{K}} P(\mathbf{y} | \mathbf{x}, k) P(k) \nonumber\\
    & \le -\sum_{k \in \mathcal{K}} \log_2 P(\mathbf{y} | \mathbf{x}, k) P(k) \label{eq:knnloss_2_1}\\
    & \le -\sum_{k \in \mathcal{K}} \log_2 P(\mathbf{y} | \mathbf{x}, k)\label{eq:knnloss_2_2}
\end{align}
\vspace{-1mm}

To proceed with Eq.~\ref{eq:knnloss_2_2}, we expand $P(\mathbf{y} | \mathbf{x}, k)$ as:
{\small
\begin{align}
    P(\mathbf{y} | \mathbf{x}, k)
    & = \frac{P(\mathbf{y},\mathbf{x},k)}{P(\mathbf{x})P(k)}
      = \frac{1}{P(\mathbf{x})P(k)} \sum_{\phi \in \Phi} P(\mathbf{y}, \mathbf{x}, k, \phi) \nonumber\\
    & = \frac{1}{P(\mathbf{x})P(k)} \sum_{\phi \in \Phi} P(\mathbf{x}) P(k) P(\phi | \mathbf{x}, k) P(\mathbf{y} | \mathbf{x}, k, \phi) \nonumber\\
    & = \sum_{\phi \in \Phi} P(\phi | \mathbf{x}, k) P(\mathbf{y} | \mathbf{x}, k, \phi) \nonumber\\
    & = P(\bar{\phi}| \mathbf{x}, k) P(\mathbf{y} | \mathbf{x}, k, \bar{\phi})
      + \hspace{-3mm}\sum_{\phi \in \Phi \setminus \bar{\phi}} P(\phi | \mathbf{x}, k) P(\mathbf{y} | \mathbf{x}, k, \phi) \nonumber\\
    & = P(\mathbf{y} | \mathbf{x}, k, \bar{\phi})\label{eq:knnloss_3}
\end{align}
}

\noindent Here, the first step uses Bayes' rule and the independence between $\mathbf{x}$ and $k$. The subsequent steps use the dependency between $\mathbf{y}$, $\mathbf{x}$, $k$, and $\phi$ shown in Figure~\ref{fig:probability}. For the final step, we reduce the summation over all possible $\phi$ to the sorted case $\bar{\phi}$, where all elements in $\phi$ are sorted in descending order. Thus, $P(\bar{\phi} | \mathbf{x}, k)=1$ and the other orders have probability $0$, since \model\ only examines sorted $\phi$.

By substituting Eq.~\ref{eq:knnloss_3} into Eq.~\ref{eq:knnloss_2}, $\mathcal{L}_{knn}^{+}$ becomes:
\begin{equation}\label{eq:knnloss_4}
    \mathcal{L}_{knn}^{+}(\mathbf{y}, \mathbf{x})  = -\sum_{k \in \mathcal{K}} \log_2  P(\mathbf{y} | \mathbf{x}, k, \bar{\phi})
\end{equation}

We now formulate the likelihood $P(\mathbf{y} | \mathbf{x}, k, \phi)$ using the  Bradley-Terry model with a reward function $w(\mathbf{y})$~\cite{bradleyterry}.
The Bradley-Terry model estimates the outcome of pairwise comparisons, determining the probability that one item is preferred over another. Formally,
\begin{equation}\label{eq:knnloss_bt}
P(y_i > y_j | x_i, x_j, k, \phi) = \sigma(x_i - x_j),
\end{equation}

\noindent where $\sigma$ is the \emph{sigmoid} function, and $x_i$ (or $y_i$) denotes the $i$-th element of $x$ (or $y$). The Bradley-Terry model matches the $k$NN objective by comparing top-$k$ items ($i \in [1,k]$) against non-top-$k$ items ($j \in (k,N]$) using predicted similarity scores.

Substituting Eq.~\ref{eq:knnloss_bt} into Eq.~\ref{eq:knnloss_4} and expanding the pairwise comparisons gives:
{\small
\begin{align}
    \mathcal{L}_{knn}^{+}(\mathbf{y}, \mathbf{x})
    & = -\sum_{k \in \mathcal{K}} \log_{2} \prod_{i = 1}^{k} \prod_{j = k+1}^{N} P(y_i > y_j | x_i, x_j, k, \bar{\phi})^{w(\mathbf{y})} \nonumber\\
    & = -\sum_{k \in \mathcal{K}} \sum_{i = 1}^{k} \sum_{j = k+1}^{N} \log_{2} P(y_i > y_j | x_i, x_j, k, \bar{\phi})^{w(\mathbf{y})} \nonumber\\
    & \leqslant -N\sum_{y_i > y_j} \log_{2} P(y_i > y_j | x_i, x_j, \bar{\phi})^{w(\mathbf{y})} \label{eq:knnloss_5}
\end{align}
}

The term in Eq.~\ref{eq:knnloss_5}, denoted as $\mathcal{L}_{knn}^{++}$, is a scaled version of the \emph{LambdaLoss}~\cite{lambdaloss}, a loss used in learning-to-rank problems. The upper bound compares only elements ranked before and after each top-$k$ boundary, which captures the $k$NN guidance while limiting the number of comparisons to $O(N^2)$.

\paragraph{The Overall Loss Function} We implement $w(\mathbf{y})$ with the approximated Normalized Discounted Cumulative Gain~(NDCG) reward (Eq.~(16) of \cite{lambdaloss}). Combining Eq.~\ref{eq:knnloss_bt} with this reward and substituting them into Eq.~\ref{eq:knnloss_5}, we obtain
$\mathcal{L}_{knn}^{++}$ over all training samples by calculating $\mathbb{E}_{\{(x, y)\}}\mathcal{L}_{knn}^{++}(y, x)$:
\begin{equation}\label{eq:knnloss_6}
    \mathcal{L}_{knn}^{++}  =  \mathop{\mathbb{E}}_{\{(x, y)\}} -N\sum_{y_i > y_j} \delta_{i,j}(G_i - G_j) \log_{2} \sigma(x_i - x_j)
\end{equation}

\noindent The terms are defined as:
{\small
\[
\begin{aligned}
G_i &= \frac{2^{y_i}-1}{\mathrm{maxDCG}},\quad
\mathrm{maxDCG}=\sum_i\frac{2^{y_i}-1}{\log_2(i+1)},\\
\delta_{i,j} &= \frac{1}{\log_2(j-i+1)}-\frac{1}{\log_2(j-i+2)}.
\end{aligned}
\]
}

Our \emph{overall loss function} $\mathcal{L}$ combines the $k$NN-guided loss with a weighted version of the MSE loss, denoted as $\mathcal{L}_{mse}$:
{\small
\begin{equation}\label{eq:mse_loss}
\mathcal{L}_{mse} = \displaystyle \mathop{\mathbb{E}}_{ T_i, T_j \in \mathcal{D}} w_{i,j} \cdot \big(f(T_i, T_j) - f'(T_i, T_j)\big)^2
\end{equation}
}

We set $w_{i,j}=f(T_i, T_j)$ so that MSE emphasizes similar pairs, and balance the two loss terms with $\lambda \in (0,1)$:
\begin{equation}\label{eq:loss}
    \mathcal{L} = \lambda\mathcal{L}_{mse} + (1-\lambda)\mathcal{L}_{knn}^{++}
\end{equation}

\subsection{Time Complexity Analysis}
\label{subsec:method_timecomplexity}

\paragraph{Trajectory Encoding}
We analyze the time complexity of \model\ for encoding a trajectory $T=[p_1,p_2,\cdots,p_n]$ into its embedding $\mathbf{h}\in\mathbb{R}^{d}$.

(1) SVEnc. First, feature preparation takes $O(n)$ time. 
Then, SVEnc applies a constant number of Conv1D and linear layers, resulting in a time complexity of $O(n\cdot d^{2})$. 
The key benefit of SVEnc is that it reduces the subsequent self-attention length from $n$ points to $m$ sub-views ($m<n$), substantially improving the efficiency of the subsequent TrajEnc module.

(2) TrajEnc. Given $\mathbf{X}$, the multi-head self-attention (Eq.~\ref{eq:selfattention}) takes $O(m^{2}\cdot d)$ time to form an $m\times m$ attention matrix capturing pairwise sub-view correlations, plus $O(m\cdot d^{2})$ time for the linear projections, yielding $O(m^{2}\cdot d+m\cdot d^{2})$ in total.

Therefore, the overall encoding time complexity of \model\ is $O(n\cdot d^{2} +m^{2}\cdot d+m\cdot d^{2})$, which is lower than applying self-attention directly on $n$ points ($O(n^{2}\cdot d)$) due to $m<n$.

\paragraph{Loss Computation} 
For training instances in a mini-batch of $N$ trajectories in Eq.~\ref{eq:knnloss_6}, 
$\mathcal{L}_{knn}^{++}$ involves at most $O(N^2)$ pairwise comparisons (see Eq.~\ref{eq:knnloss_5}), and each embedding computation takes $O(d)$ time. Therefore, the overall embedding comparison in a mini-batch takes $O(N^2 \cdot d)$ time.
In addition, the NDCG-based weights require sorting the predicted scores for each query candidate, which costs $O(N\cdot\log N)$ time per score list and $O(N^2\cdot\log N)$ time in total. The full loss computation then takes $O(N^2 \cdot d+N^2\cdot\log N)$ time.

\paragraph{Overall Time Complexity} Let $n$ and $m$ denote the typical point/sub-view lengths in a mini-batch of size $N$. The typical training time per mini-batch is dominated by encoding $N$ trajectories, i.e., $O(N\cdot(n\cdot d^2+m^{2}\cdot d+m\cdot d^{2}))$, and computing the $k$NN-guided loss, i.e.,  $O(N^{2}\cdot d+N^2\cdot\log N)$. 

\section{Experiments}\label{sec:exp}

\subsection{Experimental Settings}\label{subsec:exp:setting}

\begin{table}[bt]
\caption{Dataset Statistics.}
\centering
\adjustbox{max width=\columnwidth}{%
\footnotesize
\setlength{\tabcolsep}{5pt}
\begin{tabular}{@{}lrrr@{}}
\toprule
& \multicolumn{1}{c}{\textbf{Porto}} & \multicolumn{1}{c}{\textbf{Xian}} & \multicolumn{1}{c}{\textbf{Germany}} \\ \midrule
\#trajectories & 1,372,725 & 900,562 & 420,074 \\
Avg. \#points per traj. & 48 & 118 & 72 \\
Max. \#points per traj. & 200 & 200 & 200 \\
Avg. traj. length (km) & 6.37 & 3.25 & 252.49 \\
Max. traj. length (km) & 80.61 & 99.41 & 1.16e5 \\
Spatial area (km$^2$) & 16.0 $\times$ 20.1 & 9.9 $\times$ 9.6 & 1.0e3 $\times$ 1.2e3 \\
\bottomrule
\end{tabular}}
\vspace{-1mm}
\label{tab:apdx:dataset_statistics}
\end{table}

\paragraph{Datasets} We use three widely used large trajectory datasets: Porto~\cite{porto}, Xian~\cite{didi}, and Germany~\cite{osmplanet}. 
\textbf{Porto}~\cite{porto} contains 1.7 million taxi trajectories collected from Porto, Portugal, between July 2013 and June 2014.
\textbf{Xian}~\cite{didi} contains 2.1 million ride-hailing trajectories collected from Xi’an, China, in the first two weeks of October 2018.
\textbf{Germany}~\cite{osmplanet} contains 0.4 million trajectories from OpenStreetMap collected across Germany before 2013. 
Following previous studies~\cite{neutraj,t3s,trajgat,trajcl}, we remove consecutive, duplicated points and filter out trajectories with fewer than 20 points or more than 200 points.
We randomly sample 10,000 trajectories from each dataset, which is further split by 7:1:2 for training, validation, and test. 
Table~\ref{tab:apdx:dataset_statistics} summarizes the dataset statistics.

\paragraph{Competitors}
We compare \model\ with seven baselines:  
(1)~\textbf{NeuTraj}~\cite{neutraj} adopts a GRU with a spatial attention memory unit to encode point-based trajectories.
(2)~\textbf{T3S}~\cite{t3s} uses an LSTM and a self-attention model to encode point-based and cell-based trajectory input, respectively.
(3)~\textbf{TMN}~\cite{tmn} extends NeuTraj into a dual-branch structure to simulate the similarity computation process of two trajectories.
(4)~\textbf{TrajGAT}~\cite{trajgat} is a graph-attention-based model. It also adopts a quadtree to partition the space into grid cells of different sizes.
(5)~\textbf{TrajCL}~\cite{trajcl} introduces a dual-feature self-attention to jointly learn point-based and cell-based trajectory input.
(6)~\textbf{KGTS}~\cite{kgts} creates a knowledge graph on the grid space to help learn cell embeddings and then uses a GRU to embed trajectories based on such embeddings.
(7)~\textbf{MovSemCL}~\cite{movsemcl} uses hierarchical patch-based attentions on movement-semantics features with curvature-guided contrastive learning.

\begin{table*}[t]
\caption{Overall Results (HR@10, HR@50, and R10@50). Best results are in bold, and second best results are underlined. `$\Delta$' denotes the improvement of \model\ over the best baseline. Average rank is the mean of the ranks across all result columns.} 
\centering
\vspace{-2mm}
\resizebox{\textwidth}{!}{%
\setlength{\tabcolsep}{4pt}
\begin{tabular}{@{}llccccccccccccc@{}}
\toprule
 &  & \multicolumn{3}{c}{\textbf{DTW}} & \multicolumn{3}{c}{\textbf{EDwP}} & \multicolumn{3}{c}{\textbf{Fr\'echet}} &  \multicolumn{3}{c}{\textbf{Hausdorff}} & \\ 
 \cmidrule(lr){3-5}  \cmidrule(lr){6-8}  \cmidrule(lr){9-11} \cmidrule(lr){12-14}
\multirow{-2}{*}{\textbf{Dataset}} & \multirow{-2}{*}{\textbf{Method}} & \textbf{HR@10} & \textbf{HR@50} & \textbf{R10@50} & \textbf{HR@10} & \textbf{HR@50} & \textbf{R10@50} & \textbf{HR@10} & \textbf{HR@50} & \textbf{R10@50} & \textbf{HR@10} & \textbf{HR@50} & \textbf{R10@50} & \multirow{-2}{*}{\begin{tabular}[c]{@{}c@{}}\textbf{Average}\\ \textbf{rank}\end{tabular}} \\ \midrule
 & NeuTraj & 0.445 & 0.568 & {\ul 0.824} & 0.484 & 0.601 & 0.866 & 0.539 & 0.677 & 0.930 & 0.477 & 0.650 & 0.888 & 4.2 \\
 & T3S & 0.255 & 0.406 & 0.564 & 0.423 & 0.557 & 0.795 & 0.492 & 0.659 & 0.892 & 0.478 & 0.648 & 0.886 & 5.8 \\
 & TMN & 0.228 & 0.290 & 0.520 & 0.216 & 0.285 & 0.503 & 0.215 & 0.280 & 0.542 & 0.159 & 0.217 & 0.382 & 8.0 \\
 & TrajGAT & 0.279 & 0.366 & 0.686 & 0.226 & 0.345 & 0.601 & 0.384 & 0.430 & 0.793 & {\ul 0.714} & 0.770 & {\ul 0.988} & 5.4 \\
 & TrajCL & 0.415 & 0.572 & 0.791 & 0.546 & 0.668 & 0.910 & 0.584 & 0.715 & 0.938 & 0.644 & 0.754 & 0.975 &  3.5 \\
 & KGTS & {\ul 0.464} & 0.497 & 0.784 & 0.535 & 0.501 & 0.812 & 0.343 & 0.331 & 0.618 & 0.306 & 0.315 & 0.548 & 5.8 \\
 & MovSemCL & 0.439 & {\ul 0.591} & 0.811 & {\ul 0.590} & {\ul 0.697} & {\ul 0.935} & {\ul 0.681} & {\ul 0.796} & {\ul 0.987} & 0.684 & {\ul 0.781} & 0.984 & 2.4 \\
 & \cellcolor[HTML]{EFEFEF}\model & \cellcolor[HTML]{EFEFEF}\textbf{0.765} & \cellcolor[HTML]{EFEFEF}\textbf{0.869} & \cellcolor[HTML]{EFEFEF}\textbf{0.996} & \cellcolor[HTML]{EFEFEF}\textbf{0.796} & \cellcolor[HTML]{EFEFEF}\textbf{0.881} & \cellcolor[HTML]{EFEFEF}\textbf{0.998} & \cellcolor[HTML]{EFEFEF}\textbf{0.754} & \cellcolor[HTML]{EFEFEF}\textbf{0.838} & \cellcolor[HTML]{EFEFEF}\textbf{0.996} & \cellcolor[HTML]{EFEFEF}\textbf{0.755} & \cellcolor[HTML]{EFEFEF}\textbf{0.849} & \cellcolor[HTML]{EFEFEF}\textbf{0.995} & \cellcolor[HTML]{EFEFEF}1.0 \\
\multirow{-8}{*}{\textbf{Porto}} & \multicolumn{1}{c}{$\Delta$} & \textit{+65\%} & \textit{+47\%} & \textit{+21\%} & \textit{+35\%} & \textit{+26\%} & \textit{+7\%} & \textit{+11\%} & \textit{+5\%} & \textit{+1\%} & \textit{+6\%} & \textit{+9\%} & \textit{+1\%}  \\ \midrule
 & NeuTraj & 0.547 & {\ul 0.663} & 0.882 & 0.392 & 0.504 & 0.759 & 0.591 & 0.731 & 0.955 & 0.531 & 0.715 & 0.913 & 4.5 \\
 & T3S & 0.298 & 0.474 & 0.619 & 0.334 & 0.495 & 0.705 & 0.600 & 0.735 & 0.925 & 0.461 & 0.647 & 0.844 & 6.4 \\
 & TMN & 0.385 & 0.414 & 0.686 & 0.396 & 0.417 & 0.705 & 0.389 & 0.422 & 0.729 & 0.305 & 0.365 & 0.550 & 7.4 \\
 & TrajGAT & 0.479 & 0.616 & {\ul 0.917} & 0.378 & {\ul 0.577} & 0.842 & 0.522 & 0.633 & 0.959 & {\ul 0.778} & 0.822 & {\ul 0.993} & 3.6 \\
 & TrajCL & 0.444 & 0.625 & 0.815 & 0.427 & 0.556 & 0.799 & 0.634 & 0.732 & 0.948 & 0.704 & 0.802 & 0.968 & 4.2 \\
 & KGTS & {\ul 0.553} & 0.594 & 0.868 & {\ul 0.606} & 0.574 & {\ul 0.871} & 0.458 & 0.445 & 0.754 & 0.421 & 0.457 & 0.714 & 5.0 \\
 & MovSemCL & 0.384 & 0.527 & 0.742 & 0.347 & 0.567 & 0.824 & {\ul 0.736} & {\ul 0.820} & {\ul 0.988} & 0.746 & {\ul 0.831} & 0.984 & 4.0 \\
 & \cellcolor[HTML]{EFEFEF}\model &  \cellcolor[HTML]{EFEFEF}\textbf{0.754} & \cellcolor[HTML]{EFEFEF}\textbf{0.883} & \cellcolor[HTML]{EFEFEF}\textbf{0.995} & \cellcolor[HTML]{EFEFEF}\textbf{0.797} & \cellcolor[HTML]{EFEFEF}\textbf{0.889} & \cellcolor[HTML]{EFEFEF}\textbf{0.996} & \cellcolor[HTML]{EFEFEF}\textbf{0.812} & \cellcolor[HTML]{EFEFEF}\textbf{0.890} & \cellcolor[HTML]{EFEFEF}\textbf{0.997} & \cellcolor[HTML]{EFEFEF}\textbf{0.813} & \cellcolor[HTML]{EFEFEF}\textbf{0.892} & \cellcolor[HTML]{EFEFEF}\textbf{0.997} & \cellcolor[HTML]{EFEFEF}1.0 \\
\multirow{-8}{*}{\textbf{Xian}} & \multicolumn{1}{c}{$\Delta$} & \textit{+36\%} & \textit{+33\%} & \textit{+8\%} & \textit{+31\%} & \textit{+54\%} & \textit{+14\%} & \textit{+10\%} & \textit{+8\%} & \textit{+1\%} & \textit{+4\%} & \textit{+7\%} & \textit{+0.4\%} & \multicolumn{1}{l}{} \\ \midrule
 & NeuTraj & 0.548 & 0.605 & 0.867 & 0.557 & 0.608 & 0.855 & 0.566 & 0.657 & 0.892 & 0.556 & 0.791 & 0.935 & 4.5 \\
 & T3S & 0.444 & 0.666 & 0.854 & 0.463 & 0.728 & 0.888 & 0.536 & 0.765 & 0.944 & 0.704 & {\ul 0.869} & 0.966 & 3.9 \\
 & TMN & 0.313 & 0.432 & 0.589 & 0.332 & 0.527 & 0.661 & 0.471 & 0.569 & 0.694 & 0.515 & 0.652 & 0.817 & 6.9 \\
 & TrajGAT & 0.338 & 0.503 & 0.632 & 0.185 & 0.310 & 0.435 & 0.572 & 0.755 & 0.913 & {\ul 0.785} & 0.865 & {\ul 0.997} & 5.4 \\
 & TrajCL & {\ul 0.684} & {\ul 0.761} & {\ul 0.993} & {\ul 0.678} & {\ul 0.809} & {\ul 0.980} & {\ul 0.691} & {\ul 0.856} & {\ul 0.990} & 0.644 & 0.784 & 0.928 & 2.9 \\
 & KGTS & 0.454 & 0.497 & 0.728 & 0.465 & 0.496 & 0.729 & 0.381 & 0.362 & 0.598 & 0.312 & 0.338 & 0.482 & 6.6 \\
 & MovSemCL & 0.390 & 0.543 & 0.671 & 0.280 & 0.463 & 0.611 & 0.649 & 0.828 & 0.962 & 0.740 & 0.858 & 0.967 & 4.8 \\
 & \cellcolor[HTML]{EFEFEF}\model & \cellcolor[HTML]{EFEFEF}\textbf{0.717} & \cellcolor[HTML]{EFEFEF}\textbf{0.778} & \cellcolor[HTML]{EFEFEF}\textbf{0.999} & \cellcolor[HTML]{EFEFEF}\textbf{0.822} & \cellcolor[HTML]{EFEFEF}\textbf{0.933} & \cellcolor[HTML]{EFEFEF}\textbf{0.997} & \cellcolor[HTML]{EFEFEF}\textbf{0.869} & \cellcolor[HTML]{EFEFEF}\textbf{0.932} & \cellcolor[HTML]{EFEFEF}\textbf{0.998} & \cellcolor[HTML]{EFEFEF}\textbf{0.857} & \cellcolor[HTML]{EFEFEF}\textbf{0.930} & \cellcolor[HTML]{EFEFEF}\textbf{0.998} & \cellcolor[HTML]{EFEFEF}1.0 \\
\multirow{-8}{*}{\textbf{Germany}} & \multicolumn{1}{c}{$\Delta$} & \textit{+5\%} & \textit{+2\%} & \textit{+1\%} & \textit{+21\%} & \textit{+15\%} & \textit{+2\%} & \textit{+26\%} & \textit{+9\%} & \textit{+1\%} & \textit{+9\%} & \textit{+7\%} & \textit{+0.1\%} & \\
\bottomrule
\end{tabular}%
}
\label{tab:exp:overall}
\end{table*}

\paragraph{Implementation Details}
We set the embedding dimensionality $d$ to 128. Parameter $\lambda$ in the loss function (Eq.~\eqref{eq:loss}) is set to 0.2. The trajectory encoder uses a one-layer self-attention.
We optimize \model\ by Adam~\cite{adam} with a maximum of 40 epochs and a batch size of 128. The early stop patience is set to 10.
The learning rate is initialized to 0.002 and decayed by 50\% after every 15 epochs.
For the baseline models, we use their released code and default settings, except for T3S which we implement following its proposal, as there is no released code. The side length of grid cells is set as 100 meters for all models.
We repeat each experiment three times and report the average results.
All experiments are run on a machine with an Intel Xeon Gold 6326 CPU, an NVIDIA A100 80 GB GPU and 64 GB RAM.
The source code is available at \url{https://github.com/changyanchuan/TSMini}.

\paragraph{Evaluation Metrics}
Following previous studies~\cite{neutraj,t3s,trajgat}, we report hit ratios \textbf{HR@10} and \textbf{HR@50} and recall \textbf{R10@50}. HR@$x$ is the overlapping ratio between the ground-truth top-$x$ results and the predicted top-$x$ trajectories in test sets.
R10@50 is the recall of the ground-truth top-10 results in the predicted top-50 trajectories.

\subsection{Overall Results}
\label{subsec:exp:overall}

We follow the evaluation steps used by the competitors.
For each experiment, we: (1)~train the models to approximate a non-learned measure (e.g., Fr\'echet), (2)~compute trajectory embeddings for a test set, (3)~compute the similarity between any two trajectories in the test set based on the embeddings, (4)~rank the trajectory pairs for each trajectory in the test set using the embedding-based similarity values and the non-learned measure (i.e., the ground truth), respectively, and (5) compute evaluation metrics by comparing the embedding similarity-based ranked trajectory list with that of the ground-truth list of each test trajectory.

\begin{table*}[h]
\caption{Ablation Study Results.}
\centering
\vspace{-2mm}
\resizebox{\textwidth}{!}{%
\setlength{\tabcolsep}{4pt}
\begin{tabular}{@{}llcccccccccccc@{}}
\toprule
 &  & \multicolumn{3}{c}{\textbf{DTW}} & \multicolumn{3}{c}{\textbf{EDwP}} & \multicolumn{3}{c}{\textbf{Fr\'echet}} & \multicolumn{3}{c}{\textbf{Hausdorff}} \\
\cmidrule(lr){3-5} \cmidrule(lr){6-8} \cmidrule(lr){9-11} \cmidrule(lr){12-14}
\multirow{-2}{*}{\textbf{Dataset}} & \multirow{-2}{*}{\textbf{\begin{tabular}[c]{@{}l@{}}Model\\ variant\end{tabular}}} & \textbf{HR@10} & \textbf{HR@50} & \textbf{R10@50} & \textbf{HR@10} & \textbf{HR@50} & \textbf{R10@50} & \textbf{HR@10} & \textbf{HR@50} & \textbf{R10@50} & \textbf{HR@10} & \textbf{HR@50} & \textbf{R10@50} \\ \midrule
 & \cellcolor[HTML]{EFEFEF}\model & \cellcolor[HTML]{EFEFEF}\textbf{0.765} & \cellcolor[HTML]{EFEFEF}\textbf{0.869} & \cellcolor[HTML]{EFEFEF}\textbf{0.996} & \cellcolor[HTML]{EFEFEF}\textbf{0.796} & \cellcolor[HTML]{EFEFEF}\textbf{0.881} & \cellcolor[HTML]{EFEFEF}\textbf{0.998} & \cellcolor[HTML]{EFEFEF}\textbf{0.754} & \cellcolor[HTML]{EFEFEF}\textbf{0.838} & \cellcolor[HTML]{EFEFEF}\textbf{0.996} & \cellcolor[HTML]{EFEFEF}\textbf{0.755} & \cellcolor[HTML]{EFEFEF}\textbf{0.849} & \cellcolor[HTML]{EFEFEF}\textbf{0.995} \\
 & \model-w/o-S & {\ul 0.589} & {\ul 0.725} & {\ul 0.936} & {\ul 0.540} & {\ul 0.680} & {\ul 0.901} & 0.381 & 0.554 & 0.776 & 0.321 & 0.499 & 0.71 \\
 & \model-w/o-K & 0.436 & 0.586 & 0.802 & 0.484 & 0.617 & 0.865 & {\ul 0.656} & {\ul 0.771} & {\ul 0.983} & {\ul 0.662} & {\ul 0.766} & {\ul 0.981} \\
\multirow{-4}{*}{\textbf{Porto}} & \model-w/o-SK & 0.131 & 0.262 & 0.382 & 0.118 & 0.222 & 0.332 & 0.282 & 0.455 & 0.654 & 0.167 & 0.333 & 0.464 \\ \midrule
 & \cellcolor[HTML]{EFEFEF}\model & \cellcolor[HTML]{EFEFEF}\textbf{0.754} & \cellcolor[HTML]{EFEFEF}\textbf{0.883} & \cellcolor[HTML]{EFEFEF}\textbf{0.995} & \cellcolor[HTML]{EFEFEF}\textbf{0.797} & \cellcolor[HTML]{EFEFEF}\textbf{0.889} & \cellcolor[HTML]{EFEFEF}\textbf{0.996} & \cellcolor[HTML]{EFEFEF}\textbf{0.812} & \cellcolor[HTML]{EFEFEF}\textbf{0.890} & \cellcolor[HTML]{EFEFEF}\textbf{0.997} & \cellcolor[HTML]{EFEFEF}\textbf{0.813} & \cellcolor[HTML]{EFEFEF}\textbf{0.892} & \cellcolor[HTML]{EFEFEF}\textbf{0.997} \\
 & \model-w/o-S & {\ul 0.635} & {\ul 0.778} & {\ul 0.954} & {\ul 0.600} & {\ul 0.741} & {\ul 0.931} & 0.512 & 0.669 & 0.860 & 0.518 & 0.690 & 0.878 \\
 & \model-w/o-K & 0.339 & 0.532 & 0.670 & 0.344 & 0.570 & 0.814 & {\ul 0.765} & {\ul 0.835} & {\ul 0.993} & {\ul 0.710} & {\ul 0.816} & {\ul 0.983} \\
\multirow{-4}{*}{\textbf{Xian}} & \model-w/o-SK & 0.255 & 0.468 & 0.611 & 0.120 & 0.270 & 0.381 & 0.390 & 0.570 & 0.771 & 0.321 & 0.521 & 0.702 \\ \midrule
 & \cellcolor[HTML]{EFEFEF}\model & \cellcolor[HTML]{EFEFEF}\textbf{0.717} & \cellcolor[HTML]{EFEFEF}\textbf{0.778} & \cellcolor[HTML]{EFEFEF}\textbf{0.999} & \cellcolor[HTML]{EFEFEF}\textbf{0.822} & \cellcolor[HTML]{EFEFEF}\textbf{0.933} & \cellcolor[HTML]{EFEFEF}\textbf{0.997} & \cellcolor[HTML]{EFEFEF}\textbf{0.869} & \cellcolor[HTML]{EFEFEF}\textbf{0.932} & \cellcolor[HTML]{EFEFEF}\textbf{0.998} & \cellcolor[HTML]{EFEFEF}\textbf{0.857} & \cellcolor[HTML]{EFEFEF}\textbf{0.930} & \cellcolor[HTML]{EFEFEF}\textbf{0.998} \\
 & \model-w/o-S & {\ul 0.578} & {\ul 0.746} & {\ul 0.938} & {\ul 0.585} & {\ul 0.813} & {\ul 0.963} & {\ul 0.700} & {\ul 0.855} & {\ul 0.996} & {\ul 0.775} & {\ul 0.886} & {\ul 0.997} \\
 & \model-w/o-K & 0.317 & 0.531 & 0.665 & 0.344 & 0.551 & 0.701 & 0.388 & 0.583 & 0.724 & 0.324 & 0.507 & 0.695 \\
\multirow{-4}{*}{\textbf{Germany}} & \model-w/o-SK & 0.133 & 0.201 & 0.282 & 0.129 & 0.213 & 0.284 & 0.167 & 0.312 & 0.541 & 0.112 & 0.196 & 0.249 \\
\bottomrule
\end{tabular}%
}
\label{tab:apdx:ablation}
\vspace{-1mm}
\end{table*}

The overall results are shown in Table~\ref{tab:exp:overall}.
We see that \model\ is consistently the most accurate, achieving improvements of 22\%, 19\%, and 5\% on average over the three datasets and four non-learned measures, compared with the respective strongest baseline in HR@10, HR@50, and R10@50, respectively (15\% overall).
We made the following observations:

(1)~\model\ achieves high R10@50 values (typically \textgreater 0.99), indicating that genuinely similar trajectories are ranked near the top of the predicted list.

(2)~Among the baselines, MovSemCL and TrajCL are the overall best ones across the settings. Despite varying baseline behaviors across different datasets and distance measures, \model\ remains consistently strong because the $k$NN-guided loss directly optimizes relative similarity rankings.

(3)~KGTS frequently shows HR@50 lower (i.e., worse) than HR@10. This is because KGTS is trained without using any non-learned measure. The unlabeled data helps little in approximating specific non-learned measures.

\subsection{Ablation Study}\label{subsec:exp:ablation}
\paragraph{Model Variants and Results}
We compare \model\ with three model variants:
\textbf{\model-w/o-S} replaces the sub-view encoder SVEnc with the cell encoder used in T3S and TrajCL;
\textbf{\model-w/o-K} removes the $k$NN-guided loss from Eq.~\eqref{eq:loss}; and
\textbf{\model-w/o-SK} uses the cell encoder and the MSE loss.

We repeat the experiments as above.
As Table~\ref{tab:apdx:ablation} shows, both the $k$NN-guided loss and SVEnc
are important to the overall model performance. They improve the hit ratios by 88\% (\model-w/o-K vs. \model) and 41\% (\model-w/o-S vs. \model) in HR@10 on average, respectively.
In most cases, the $k$NN-guided loss plays a more critical role, especially in learning DTW and EDwP, whereas the sub-view modeling is more effective in learning Fr\'echet.
\paragraph{Impact of Self-Attention Modules}
We further investigate the impact of self-attention modules.
\model\ uses a Transformer encoder layer with RMSNorm, RoPE, and SwiGLU.
We compare it with two model variants:~\textbf{\model-GQA}, which adds GQA-style parameter sharing, and~\textbf{\model-vanillaSA}, which uses vanilla Transformer self-attention~\cite{transformer}.
We repeat the above experiments and report the HR@10 results. Similar result patterns are seen with the other metrics.
\begin{table}[t]
\caption{Effectiveness of the Self-attention Module in the Trajectory Encoder of \model\ (HR@10).}
\small
\centering
\adjustbox{max width=\columnwidth}{%
\setlength{\tabcolsep}{5pt}
\begin{tabular}{@{}llcccc@{}}
\toprule
\textbf{Dataset} & \textbf{Model variant} & \textbf{DTW} & \textbf{EDwP} & \textbf{Fr\'echet} & \textbf{Hausdorff} \\ \midrule
\multirow{3}{*}{\textbf{Porto}} & \cellcolor[HTML]{EFEFEF}\model\ & \cellcolor[HTML]{EFEFEF}\textbf{0.765} & \cellcolor[HTML]{EFEFEF}\textbf{0.796} & \cellcolor[HTML]{EFEFEF}\textbf{0.754} & \cellcolor[HTML]{EFEFEF}\textbf{0.755} \\
 & \model-GQA & {\ul 0.761} & {\ul 0.789} & {\ul 0.752} & {\ul 0.739} \\
 & \model-vanillaSA & 0.751 & 0.793 & 0.740 & 0.725 \\ \midrule
\multirow{3}{*}{\textbf{Xian}} & \cellcolor[HTML]{EFEFEF}\model\ & \cellcolor[HTML]{EFEFEF}\textbf{0.754} & \cellcolor[HTML]{EFEFEF}\textbf{0.813} & \cellcolor[HTML]{EFEFEF}\textbf{0.812} & \cellcolor[HTML]{EFEFEF}\textbf{0.812} \\
 & \model-GQA & 0.751 & {\ul 0.788} & {\ul 0.807} & {\ul 0.797} \\
 & \model-vanillaSA & {\ul 0.751} & 0.783 & 0.799 & 0.798 \\ \midrule
\multirow{3}{*}{\textbf{Germany}} & \cellcolor[HTML]{EFEFEF}\model\ & \cellcolor[HTML]{EFEFEF}\textbf{0.717} & \cellcolor[HTML]{EFEFEF}\textbf{0.822} & \cellcolor[HTML]{EFEFEF}\textbf{0.869} & \cellcolor[HTML]{EFEFEF}\textbf{0.857} \\
 & \model-GQA & 0.683 & 0.787 & 0.826 & 0.815 \\
 & \model-vanillaSA & {\ul 0.704} & {\ul 0.811} & {\ul 0.858} & {\ul 0.835} \\
\bottomrule
\end{tabular}}
\label{tab:apdx:self-attention}
\vspace{-2mm}
\end{table}

Table~\ref{tab:apdx:self-attention} presents the results.
\model\ outperforms \model-GQA and \model-vanillaSA consistently, showing that GQA-style parameter sharing is not beneficial for trajectory encoding. The performance of \model-GQA shows an average gap of 2.5\% to \model, suggesting that sharing attention parameters slightly harms the encoding quality for our task. Moreover, as shown in the efficiency study (Section~\ref{sec:exp:efficiency}), \model\ is already one of the fastest models at both training and inference, eliminating the need for GQA to trade accuracy for speed.
\model\ also outperforms \model-vanillaSA by 1.9\% on average.

These results show that the difference in performance among \model, \model-GQA, and \model-vanillaSA is relatively small. They confirm that the main performance improvements of \model\ over the existing models stem from the sub-view encoder and $k$NN-guided loss, rather than from specific Transformer-layer design choices.

\begin{table}[t]
\caption{Embedding Quality of \model\ with and without Sub-View Encoder on Xian --  Rand Index (RI) of Trajectory Clustering.}
\footnotesize
\centering
\adjustbox{max width=\columnwidth}{%
\setlength{\tabcolsep}{6pt}
\begin{tabular}{@{}lcccc@{}}
\toprule
 & \textbf{DTW} & \textbf{EDwP} & \textbf{Fr\'echet} & \textbf{Hausdorff} \\ \midrule
\cellcolor[HTML]{EFEFEF}\model & \cellcolor[HTML]{EFEFEF}\textbf{0.919} & \cellcolor[HTML]{EFEFEF}\textbf{0.954} & \cellcolor[HTML]{EFEFEF}\textbf{0.939} & \cellcolor[HTML]{EFEFEF}\textbf{0.966} \\
\model-w/o-S & 0.903 & 0.935 & 0.929 & 0.932 \\
\bottomrule
\end{tabular}}
\label{xtab:ri}
\vspace{-2mm}
\end{table}

\paragraph{Embedding Quality of the Sub-View Encoder}
Apart from the ablation study on SVEnc above, we further conduct experiments to investigate the effectiveness of SVEnc in capturing movement patterns.
We evaluate the clustering accuracy of \model\ (with sub-view-based inputs) and \model\-w/o-S (using the cell-based inputs as in T3S and TrajCL) based on the generated trajectory embeddings, as this can indicate how SVEnc learns trajectory representations between similar trajectories.
We follow the experimental settings used in a recent study~\cite{trajsimi_survey}, applying the $k$-medoids algorithm ($k$=10) and measuring cluster accuracy using the Rand Index (RI). RI (higher the better) measures the percentage of ground-truth similar trajectory pairs (based on non-learned measures) being correctly assigned to the same cluster. We report the results on Xian in Table~\ref{xtab:ri} (similar results are observed on the other datasets and are omitted for conciseness).
We see that \model\ achieves higher RI consistently than the model variant using cell-based inputs, verifying that SVEnc helps \model\ produce more accurate trajectory representations.

\begin{figure}[!th]
    \centering
    \vspace{-2mm}
    \includegraphics[width=1.0\columnwidth]{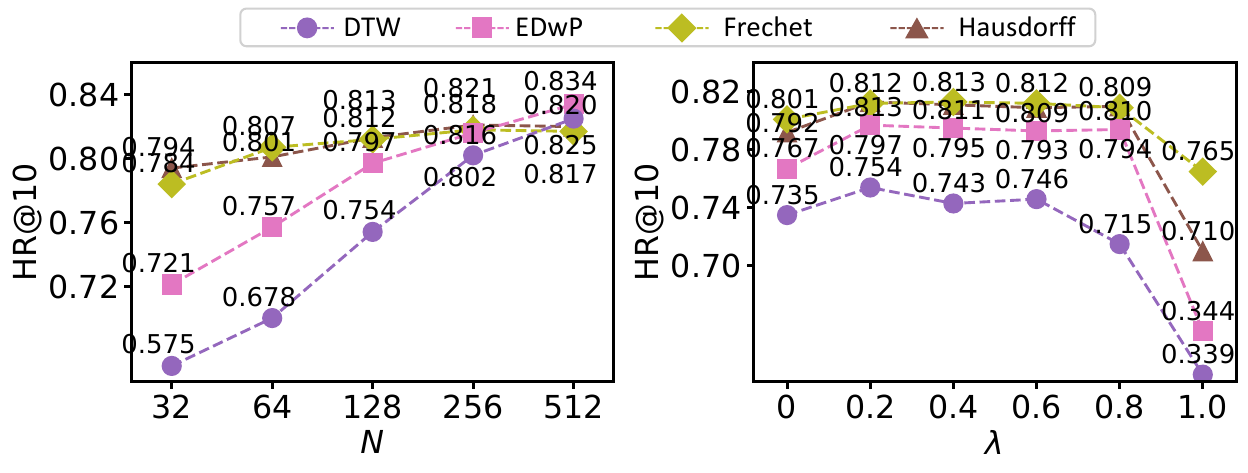}
    \vspace{-5mm}
    \caption{Parameter study on Xian. The left and right panels report HR@10 when varying the batch size $N$ and $\lambda$, respectively.}\label{fig:parameters}
    \vspace{-2mm}
\end{figure}

\subsection{Parameter Study}
\label{subsec:exp:parameter}
Using Xian as a representative dataset, we investigate how the batch size $N$ and the hyper-parameter $\lambda$ impact \model.

\paragraph{Impact of Batch Size}
In practice, we compute the $k$NN-guided loss within each mini-batch. Thus, the $N$ trajectories in Eq.~\eqref{eq:knnloss_6} correspond to the batch size in this experiment.
The left subfigure of Figure~\ref{fig:parameters} shows the results on varying $N$. The HR@10 results generally increase with $N$. This is expected, since a larger batch size (i.e., a larger $N$ in Eq.~\eqref{eq:knnloss_6}) allows \model\ to learn more relative similarity values from the $k$NN-guided loss.
Table~\ref{tab:batch_training_time} reports the average training time over four measures and two random seeds on Xian.
Larger batches reduce the number of parameter-update steps per epoch and improve GPU utilization, which shortens the total training time.
We use a batch size of 128 by default to keep inline with the baselines for ease of comparison, although a larger batch size could further improve our model accuracy.

\begin{table}[t]
\caption{Average Training Time with Different Batch Sizes on Xian.}
\centering
\vspace{-2mm}
\adjustbox{max width=\columnwidth}{%
\footnotesize
\setlength{\tabcolsep}{6pt}
\begin{tabular}{@{}lccccc@{}}
\toprule
\textbf{Batch size $N$} & 32 & 64 & 128 & 256 & 512 \\ \midrule
\textbf{Avg. TrT (h)} & 2.40 & 1.43 & 0.76 & 0.44 & 0.38 \\
\bottomrule
\end{tabular}}
\label{tab:batch_training_time}
\vspace{-1mm}
\end{table}

\paragraph{Impact of $\lambda$}
The right subfigure of Figure~\ref{fig:parameters} shows the results on varying $\lambda$ in Eq.~\eqref{eq:loss}. Except when $\lambda$ is 0 or 1, the value of $\lambda$ has a limited impact on \model.
Such results indicate that both loss terms contribute to the accuracy of \model, while using the $k$NN-guided loss alone ($\lambda=0$) is more effective than using the MSE loss ($\lambda=1$).

\begin{figure}[t]
 \vspace{-2mm}
    \hspace*{2mm}
    \includegraphics[width=0.46\textwidth]{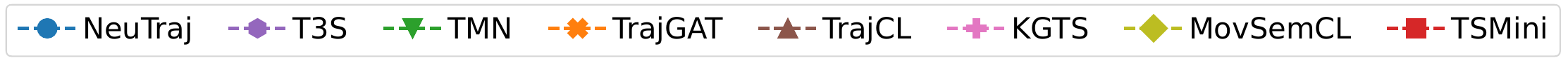} \\
     \vspace{-5mm}

     \hspace*{-2mm}
     \subfloat[{Xian}~\label{fig:exp_num_points_xian_hr10}]{
        \includegraphics[width=0.22\textwidth]{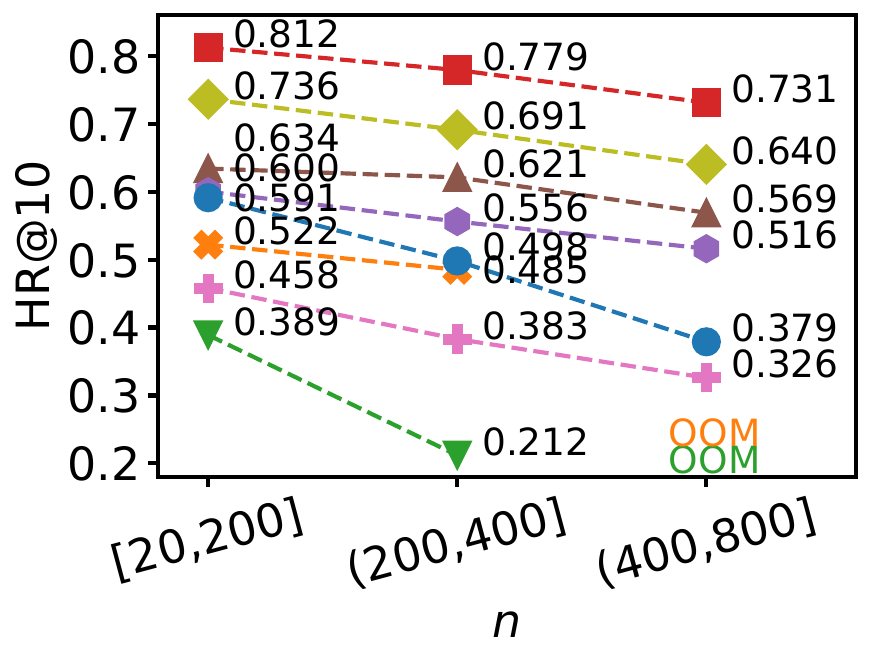}
    }\hspace*{-2mm}
    \subfloat[{Germany}~\label{fig:exp_num_points_germany_hr10}]{
        \includegraphics[width=0.22\textwidth]{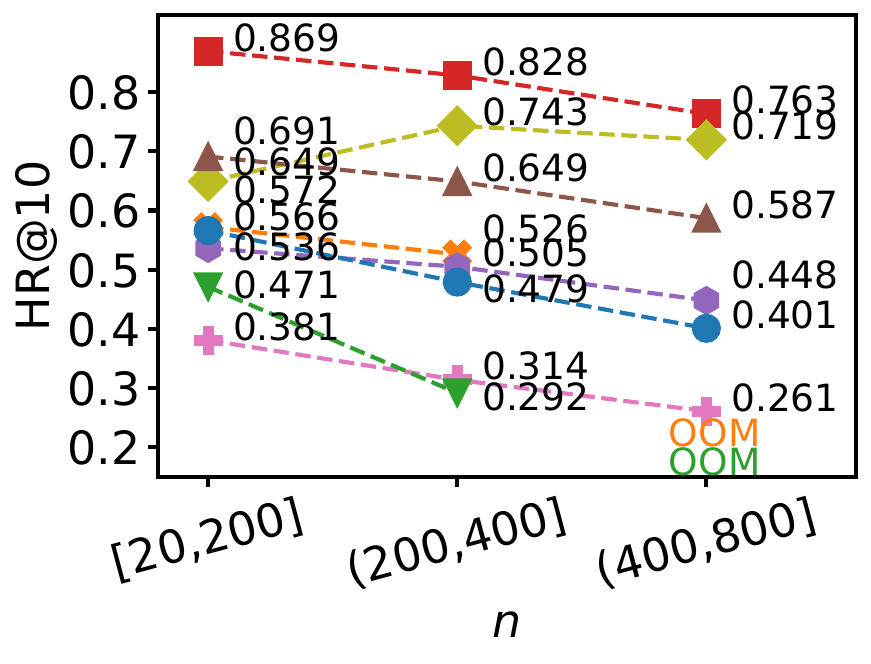}
    }
    \caption{Varying the number of points on trajectories to approximate Fr\'echet (OOM: out-of-memory error).}\label{fig:exp_num_points}
\end{figure}

\subsection{Impact of Data Scale}
\paragraph{Varying the Number of Points on Trajectories ($n$)}
We investigate the effectiveness of TSMini on longer trajectories.
We randomly sample another two datasets of 10,000 trajectories from Xian and Germany (Porto does not have as many long trajectories), with 200 to 400, and 400 to 800 points on trajectories, respectively.
We report HR@10 on approximating Fr\'echet. Similar result patterns are observed on the other similarity measures and performance metrics, which are omitted for conciseness. Porto does not have as many long trajectories and is not used for this set of experiments.

Figure~\ref{fig:exp_num_points} shows that, overall, the accuracy of all methods decreases as the length of trajectories increases, since longer trajectories bring challenges.
\model\ consistently outperforms all baselines, especially achieving higher HR@10 when $n=(400,800]$ compared to other methods at $n=[20,200]$. This highlights the superior capability of~\model\ on learning long trajectories.
TMN and TrajGAT incur out-of-memory (OOM) errors when $n=(400,800]$. This is because TMN requires computing the correlation between all point pairs in two trajectories, while TrajGAT constructs large graphs to represent long trajectories, both of which are memory-intensive. We observe an increase in HR@10 for MovSemCL on Germany when $n$ grows from $[20,200]$ to $(400,800]$. We conjecture that MovSemCL's fixed patch size design is suboptimal for shorter trajectories on this dataset, but becomes more suitable as trajectories grow longer and are segmented into more patches, enabling better capture of local patterns.

\paragraph{Impact of Training Set Size}\label{app:training_set_size}
We investigate the impact of training set size on model accuracy. Recall that, by default, each dataset in the experiments contains 7,000, 1,000 and 2,000 trajectories for training, validation and testing, respectively. We increase the training set sizes to 14,000 and 28,000 while keeping the validation and test sets unchanged. These expanded datasets are referred to with ``-2x'' and ``-4x'' in Table~\ref{xtab:vary_train_set_size}, respectively.
We compare \model\ with TrajCL and MovSemCL, two of the strongest baselines, for approximating Fr\'echet on Xian.
The GT time column reports the one-off cost to compute ground-truth similarity labels before model training.

As Table~\ref{xtab:vary_train_set_size} shows, \model\ consistently outperforms TrajCL, while additional training data does not lead to significant accuracy improvements. Notably, \model\ on Xian achieves higher accuracy than TrajCL and MovSemCL on Xian-4x, even though Xian-4x uses a larger training set. Such results indicate that \model\ remains effective even with a smaller training set, making it highly practical.
The GT time results further show that conventional similarity computation is costly, which motivates learning a reusable neural approximator for efficient subsequent similarity queries.

\begin{table}[t]
\caption{Impact of Training Set Size.}
\centering
\adjustbox{max width=\columnwidth}{%
\footnotesize
\setlength{\tabcolsep}{5pt}
\begin{tabular}{@{}lclccc@{}}
\toprule
\textbf{Dataset} & \textbf{GT time} & \textbf{Method} & \textbf{HR@10} & \textbf{HR@50} & \textbf{R10@50} \\ \midrule
\multirow{3}{*}{\textbf{Xian}} & \multirow{3}{*}{5,303s} & TrajCL & 0.634 & 0.732 & 0.948 \\
 & & MovSemCL & 0.736 & 0.820 & 0.988 \\
 & & \cellcolor[HTML]{EFEFEF}\model & \cellcolor[HTML]{EFEFEF}\textbf{0.812} & \cellcolor[HTML]{EFEFEF}\textbf{0.890} & \cellcolor[HTML]{EFEFEF}\textbf{0.997} \\ \midrule
\multirow{3}{*}{\textbf{Xian-2x}} & \multirow{3}{*}{10,636s} & TrajCL & 0.726 & 0.811 & 0.987 \\
 & & MovSemCL & 0.768 & 0.836 & 0.994 \\
 & & \cellcolor[HTML]{EFEFEF}\model & \cellcolor[HTML]{EFEFEF}\textbf{0.822} & \cellcolor[HTML]{EFEFEF}\textbf{0.899} & \cellcolor[HTML]{EFEFEF}\textbf{0.999} \\ \midrule
\multirow{3}{*}{\textbf{Xian-4x}} & \multirow{3}{*}{52,805s} & TrajCL & 0.729 & 0.816 & 0.990 \\
 & & MovSemCL & 0.767 & 0.837 & 0.996 \\
 & & \cellcolor[HTML]{EFEFEF}\model & \cellcolor[HTML]{EFEFEF}\textbf{0.824} & \cellcolor[HTML]{EFEFEF}\textbf{0.900} & \cellcolor[HTML]{EFEFEF}\textbf{0.999} \\
\bottomrule
\end{tabular}}
\vspace{-2mm}
\label{xtab:vary_train_set_size}
\end{table}

\subsection{Robustness to Data Quality and Distribution Shift}
\paragraph{Evaluation on Low Quality Data}
To further examine the robustness of~\model, we evaluate it  on intentionally degraded, low-quality data.
Following the setup of the TrajCL paper~\cite{trajcl}, we simulate noisy or incomplete trajectories by randomly masking and shifting points in the test set trajectories.
Specifically, we randomly mask 40\% and 80\% of the points in each trajectory (denoted as $M_{40\%}$ and $M_{80\%}$, respectively), and/or apply random spatial shifts of 100 or 500 meters to these points (denoted as $S_{100}$ and $S_{500}$, respectively).
We report HR@10 for learning Fr\'echet distance on the Xian dataset.

As shown in Table~\ref{xtab:low_quality_data}, \model\ outperforms both TrajCL and MovSemCL in most scenarios, with average improvements of 34.3\% and 6.7\% respectively, demonstrating that its continuous space representation enhances generalization compared to grid-based discrete methods. MovSemCL exhibits an interesting robustness pattern under combined perturbations: its performance on $M_{40\%}$+$S_{100}$ (0.730) and $M_{80\%}$+$S_{500}$ (0.655) degrades less than expected from individual effects. This can be attributed to its patch-based encoding mechanism. When masking reduces trajectory length, fewer, but still semantically coherent, patches are processed, partially mitigating the impact of spatial noise on the remaining points.

\begin{table}[t]
\caption{Model Effectiveness on Low Quality Test Trajectories in HR@10.}
\centering
\adjustbox{max width=\columnwidth}{%
\footnotesize
\setlength{\tabcolsep}{3pt}
\begin{tabular}{@{}lccccccc@{}}
\toprule
 & Original & $M_{40\%}$ & $M_{80\%}$ & $S_{100}$ & $S_{500}$ & $M_{40\%}$+$S_{100}$ & $M_{80\%}$+$S_{500}$ \\ \midrule
TrajCL & 0.634 & 0.632 & 0.583 & 0.545 & 0.470 & 0.502 & 0.452 \\
MovSemCL & 0.736 & 0.735 & 0.655 & 0.716 & 0.562 & 0.730 & \textbf{0.655} \\
\cellcolor[HTML]{EFEFEF}\model & \cellcolor[HTML]{EFEFEF}\textbf{0.812} & \cellcolor[HTML]{EFEFEF}\textbf{0.799} & \cellcolor[HTML]{EFEFEF}\textbf{0.669} & \cellcolor[HTML]{EFEFEF}\textbf{0.793} & \cellcolor[HTML]{EFEFEF}\textbf{0.686} & \cellcolor[HTML]{EFEFEF}\textbf{0.783} & \cellcolor[HTML]{EFEFEF}0.559 \\
\bottomrule
\end{tabular}}
\label{xtab:low_quality_data}
\end{table}

\paragraph{Evaluation on Test Sets in Unseen Distributions}
To further assess the generalization capability of~\model, we perform cross-dataset experiments where the model is trained on one dataset and directly tested on another unseen dataset, following the setup in TrajCL~\cite{trajcl}.
This setting poses a significant challenge, as it evaluates the model on trajectory distributions that differ substantially from those observed during training.
It thus serves as a rigorous test of the model's robustness and ability to generalize beyond seen data.

Table~\ref{xtab:diff_distribution} summarizes the results, where $A \rightarrow B$ indicates that the model is trained on dataset $A$ and evaluated on dataset $B$.
As shown in the table,~\model\ consistently outperforms TrajCL and MovSemCL by an average margin of 41\% and 70\%, respectively.
Moreover, when compared to the results of querying on the original in-distribution trajectories (as reported in Table \ref{tab:exp:overall}),~\model\ experiences only a modest 24\% performance degradation, whereas TrajCL suffers a larger 36\% drop, and MovSemCL exhibits a severe 69\% degradation.
These findings confirm that~\model\ exhibits stronger generalization and transferability across diverse trajectory distributions, highlighting its robustness under domain shift scenarios.

\begin{table}[t]
\caption{Model Effectiveness on Test Sets in Unseen Distributions in HR@10.}
\centering
\adjustbox{max width=\columnwidth}{%
\footnotesize
\setlength{\tabcolsep}{3pt}
\begin{tabular}{@{}lcccccccc@{}}
\toprule
 & \multicolumn{4}{c}{\textbf{Xian $\rightarrow$ Porto}} & \multicolumn{4}{c}{\textbf{Porto $\rightarrow$ Xian}} \\ \cmidrule(lr){2-5} \cmidrule(lr){6-9}
 & DTW & EDwP & Fr\'echet & Hausdorff & DTW & EDwP & Fr\'echet & Hausdorff \\ \midrule
TrajCL & 0.230 & 0.368 & 0.548 & 0.543 & 0.261 & 0.284 & 0.254 & 0.349 \\
MovSemCL & 0.119 & 0.121 & 0.118 & 0.158 & 0.241 & 0.234 & 0.171 & 0.270 \\
\cellcolor[HTML]{EFEFEF}\model & \cellcolor[HTML]{EFEFEF}\textbf{0.509} & \cellcolor[HTML]{EFEFEF}\textbf{0.621} & \cellcolor[HTML]{EFEFEF}\textbf{0.633} & \cellcolor[HTML]{EFEFEF}\textbf{0.601} & \cellcolor[HTML]{EFEFEF}\textbf{0.494} & \cellcolor[HTML]{EFEFEF}\textbf{0.619} & \cellcolor[HTML]{EFEFEF}\textbf{0.687} & \cellcolor[HTML]{EFEFEF}\textbf{0.645} \\
\bottomrule
\end{tabular}}
\label{xtab:diff_distribution}
\end{table}

\begin{table}[t]
\caption{Space and Time Efficiency
(`TrT' and `InfT' denote training time and inference time, respectively.)}
\centering
\adjustbox{max width=\columnwidth}{%
\footnotesize
\setlength{\tabcolsep}{4pt}
\begin{tabular}{@{}lcrrrrrr@{}}
\toprule
\multirow{2}{*}{\textbf{Method}} & \multirow{2}{*}{\textbf{\#params}} & \multicolumn{2}{c}{\textbf{Porto}} & \multicolumn{2}{c}{\textbf{Xian}} & \multicolumn{2}{c}{\textbf{Germany}} \\ \cmidrule(lr){3-4} \cmidrule(lr){5-6} \cmidrule(lr){7-8}
 &  & \multicolumn{1}{c}{\textbf{TrT}} & \multicolumn{1}{c}{\textbf{InfT}} & \multicolumn{1}{c}{\textbf{TrT}} & \multicolumn{1}{c}{\textbf{InfT}} & \multicolumn{1}{c}{\textbf{TrT}} & \multicolumn{1}{c}{\textbf{InfT}} \\ \midrule
NeuTraj & \textbf{0.12M} & 5,605s & 2.47s & 5,661s & 2.96s & 6,897s & 2.62s \\
T3S & 0.20M & {\ul 2,246s} & 0.26s & {\ul 2,885s} &  0.30s & {\ul 2,373s} & 0.31s \\
TMN & {\ul 0.18M} & 3,968s & - & 6,176s & - & 4,204s & - \\
TrajGAT & 2.84M & 15,529s & 2.48s & 34,866s & 3.64s & 9,770s & 3.52s \\
TrajCL & 1.22M & 2,810s & {\ul 0.20s} & 3,006s & {\ul 0.25s} & 2,906s & {\ul 0.24s} \\
KGTS & 2.99M & 2,534s & \textbf{0.14s} & 5,600s & \textbf{0.17s} & 4,902s & \textbf{0.17s} \\
MovSemCL & 1.85M & 8,490s & 2.27s & 8,175s & 1.36s & 13,756s & 1.02s \\
\rowcolor[HTML]{EFEFEF}
\model & 0.28M & \textbf{1,917s} & 0.28s & \textbf{2,605s} & 0.34s & \textbf{1,724s} & 0.32s \\
\bottomrule
\end{tabular}}
\label{tab:exp:efficiency}
\end{table}

\subsection{Model Efficiency}\label{sec:exp:efficiency}
We further study the time and space efficiency of \model.
Table~\ref{tab:exp:efficiency} shows the results for learning to approximate Fr\'echet~(other measures have close results), where the reported time is end-to-end.
\model\ is efficient in both space and time:
(1)~It is one of the models with the fewest parameters, trailing behind NeuTraj and TMN that solely rely on RNNs, and T3S that is based on self-attention without FFNs.
These models have low accuracy as reported above.
(2)~It is the fastest in training, benefiting from the $k$NN-guided loss that leads to fast convergence.
(3)~It is among the fastest in inference, due to the highly efficient parallelization of self-attention computation.
TMN cannot be used as an encoder on its own and  is omitted.

\section{Conclusion}\label{sec:conclusion}

We proposed \model, a simple yet highly effective trajectory similarity learning model using a sub-view encoder for  trajectory modeling and a $k$NN-guided loss for model training. 
\model\ can capture trajectory movement patterns at different granularities and is optimized with the similarity between a trajectory and its $k$NNs. It thus produces trajectory representations robust to a variety of similarity measures.  
Experimental results on large trajectory datasets show that \model\ outperforms SOTA models by 15\% on average.

\section*{Acknowledgments}
This work is in part supported by the Australian Research
Council (ARC) via Discovery Projects DP230101534 and  
DP240101006. Jianzhong Qi is supported by ARC Future Fellowship FT240100170.

\bibliographystyle{IEEEtran}
\bibliography{main}

\end{document}